\documentclass{llncs}
 
\usepackage{graphicx}
\usepackage{latexsym}
\usepackage{soul}
\usepackage{comment}
\usepackage{graphicx}
\usepackage{enumitem}
\usepackage{hyperref}
\usepackage{color}
\usepackage[colorinlistoftodos]{todonotes}
\usepackage{booktabs} 
\usepackage{multirow}

\usepackage{amssymb}
\usepackage{soul}
\usepackage[normalem]{ulem}
\usepackage{xcolor}

\usepackage{placeins}

\newcommand\redsout{\bgroup\markoverwith{\textcolor{red}{\rule[0.5ex]{2pt}{0.4pt}}}\ULon}

\newcommand{\ct}{\texttt{CheckThat! }}

\newcommand{\rank}[2]{{\scriptsize \texttt{#1:#2}}}

\definecolor{darkturquoise}{rgb}{0.0, 0.81, 0.82}
\definecolor{lightblue}{rgb}{.50,.95,1}
\definecolor{tri}{rgb}{.25,.88,.82}
\definecolor{lilac}{rgb}{0.85,0.64,0.85}

\title{Overview of the CLEF--2021 CheckThat! Lab\\
on Detecting Check-Worthy Claims,\\ Previously Fact-Checked Claims, and Fake News}

\author{Preslav Nakov$^1$,  Giovanni Da San Martino$^2$, Tamer Elsayed$^3$,
\\Alberto Barr\'on-Cede\~no$^4$, Rub\'en M\'iguez$^5$, Shaden Shaar$^1$, {Firoj Alam$^1$}, \\Fatima Haouari$^3$, Maram Hasanain$^3$, Watheq Mansour$^3$,  Bayan Hamdan$^{11}$,  \\ Zien Sheikh Ali$^3$, Nikolay Babulkov$^6$, Alex Nikolov$^6$, Gautam Kishore Shahi$^7$, Julia Maria Stru\ss$^8$, Thomas Mandl$^9$, Mucahid Kutlu$^{10}$,
               Yavuz Selim Kartal$^{10}$
            \smallskip\\
    $^1$ Qatar Computing Research Institute, HBKU, Qatar \\
    $^2$ University of Padova, Italy\\
    $^3$ Qatar University, Qatar \\ 
	$^4$ DIT, Universit\`a di Bologna, Italy\\
	$^5$ Newtral Media Audiovisual, Spain\\
    $^6$ Sofia University, Bulgaria \\ 
    $^7$ University of Duisburg-Essen, Germany\\
    $^8$ University of Applied Sciences Potsdam, Germany \newline
    $^9$ University of Hildesheim, Germany \newline
    $^{10}$ TOBB University of Economics and Technology, Turkey \newline
    $^{11}$ Independent Researcher
    }

\authorrunning{Nakov et al.}
\institute{%
\email{\{pnakov,sshaar,fialam\}@hbku.edu.qa},
\email{\{telsayed,200159617,maram.hasanain,wm1900793,zs1407404\}@qu.edu.qa},
\email{dasan@math.unipd.it}, \email{a.barron@unibo.it}, \email{ruben.miguez@newtral.es},\\
\email{\{nbabulkov, alexnickolow\}@gmail.com},
\email{gautam.shahi@uni-due.de},
\email{struss@fh-potsdam.de},
\email{mandl@uni-hildesheim.de},\\
\email{\{m.kutlu, ykartal\}@etu.edu.tr},
\email{bayan.hamdan995@gmail.com}
}
\begin{document}

\maketitle

\begin{abstract}
We describe the fourth edition of the \ct Lab, part of the 2021 Conference and Labs of the Evaluation Forum (CLEF). The lab evaluates technology supporting tasks related to factuality, and covers Arabic, Bulgarian, English, Spanish, and Turkish.
Task~1 asks to predict which posts in a Twitter stream are worth fact-checking, focusing on COVID-19 and politics (in all five languages). 
Task~2 asks to determine whether a claim in a tweet can be verified using a set of previously fact-checked claims (in Arabic and English). 
Task~3 asks to predict the veracity of a news article and its topical domain (in English). 
The evaluation is based on mean average precision or precision at rank $k$ for the ranking tasks, and macro-F$_1$ for the classification tasks. This was the most popular CLEF-2021 lab in terms of team registrations: 132 teams. Nearly one-third of them participated: 15, 5, and 25 teams submitted official runs for tasks 1, 2, and 3, respectively.

\end{abstract}

\section{Introduction} 
\label{sec:introduction}

The mission of the \ct lab is to foster the development of technology to enable the (semi-)automatic verification of claims. 
Systems for claim identification and verification can be very useful as supportive technology for investigative journalism, as they could provide help and guidance, thus saving  time~\cite{RANLP2017:debates,Hassan:15,hassan2017claimbuster,RANLP2019:checkworthiness:multitask,kazemi2021tiplines}.
A system could automatically identify check-worthy claims, make sure they have not been fact-checked already by a reputable fact-checking organization, and then present them to a journalist for further analysis in a ranked list. 
Additionally, the system could identify documents that are potentially \textit{useful} for humans to perform manual fact-checking of a claim, and it could also estimate a \emph{veracity score} supported by evidence to increase the journalist's understanding and trust in the system's decision. 

\ct at CLEF 2021 is the fourth edition of the lab.
The 2018 edition~\cite{clef2018checkthat} focused on the identification and verification of claims in political debates.
The 2019 edition~\cite{ecir-checkthat:2019,clef-checkthat:2019}
featured political debates and isolated claims, in conjunction with a closed set of Web documents to retrieve evidence from.

In 2020 \cite{clef-checkthat:2020}, the focus was on social media ---in particular on \emph{Twitter}--- as information posted on this platform is not checked by an authoritative entity before posting and such posts tend to disseminate very quickly. Moreover, social media posts lack context due to their short length and conversational nature; thus, identifying a claim's context is sometimes key for effective fact-checking~\cite{cazalens2018content}.

In the 2021 edition of the \ct lab, we feature three tasks: 1.~check-worthiness estimation, 2.~detecting previously fact-checked claims, and 3.~predicting the veracity of news articles and their domain.
In these tasks, we focus on (\emph{i})~\emph{tweets}, (\emph{ii})~\emph{political debates and speeches}, and (\emph{iii})~\emph{news articles}. Moreover, besides Arabic and English, we extend our language coverage to Bulgarian, Spanish, and Turkish.
We further add a new task (task 3) on multi-class fake news detection for news articles and topical domain identification, which can help direct the article to the right fact-checking expert\cite{DBLP:conf/ecir/NakovMEBMSAHHBN21}.

\section{Previously on \ct}
\label{sec:prevct}

Three editions of the \ct lab have been held so far, and some of the tasks in the 2021 edition are reformulated from previous editions. Below, we discuss some relevant tasks from previous years.

\subsection{\ct 2020}
\paragraph{Task~1$_{2020}$.} \emph{Given a topic and a stream of potentially related tweets, rank the tweets by check-worthiness for the topic} \cite{clef-checkthat-ar:2020,clef-checkthat-en:2020}. The most successful runs adopted state-of-the-art transformer models. The top-ranked teams for the English version of this task used BERT~\cite{clef-checkthat-cheema:2020} and RoBERTa~\cite{clef-checkthat-Nikolov:2020,clef-checkthat-williams:2020}. For the Arabic version, the top systems used AraBERT~\cite{clef-checkthat-Kartal:2020,clef-checkthat-williams:2020} and the multilingual BERT~\cite{clef-checkthat-Hasanain:2020}. 

\paragraph{Task~2$_{2020}$.} \emph{Given a check-worthy claim and a dataset of verified claims, rank the verified claims, so that those that verify the input claim (or a sub-claim in it) are ranked on top of the list} \cite{clef-checkthat-en:2020}. The most effective approaches fine-tuned large-scale pre-trained transformers such as BERT and RoBERTa. In particular, the top-ranked run fine-tuned RoBERTa~\cite{clef-checkthat-Bouziane:2020}.

\paragraph{Task~4$_{2020}$.} \emph{Given a check-worthy claim on a specific topic and a set of potentially-relevant Web pages, predict the veracity of the claim}~\cite{clef-checkthat-ar:2020}. Two runs were submitted for the task~\cite{clef-checkthat-Touahri:2020}, using a scoring function that computes the degree of concordance and negation between a claim and all input text snippets for that claim.

\paragraph{Task~5$_{2020}$.} \emph{Given a political debate or a speech, segmented into sentences, together with information about who the speaker of each sentence is, prioritize the sentences for fact-checking}  \cite{clef-checkthat-en:2020}. For this task, only one out of eight runs outperformed a strong bi-LSTM baseline~\cite{clef-checkthat-Martinez-Rico:2020}.

\subsection{\ct 2019}
\paragraph{Task~1$_{2019}$.} \emph{Given a political debate, an interview, or a speech, segmented into sentences, rank the sentences by the priority with which they should be fact-checked} \cite{clef-checkthat-T1:2019}. The most successful approaches used neural networks for the classification of the individual instances. For example, Hansen et al.~\cite{T1-Hansen:2019} learned domain-specific word embeddings and syntactic dependencies and used an LSTM with a classificatiuon layer onn top of it.

\paragraph{Task~2$_{2019}$.} \emph{Given a claim and a set of potentially relevant Web pages, identify which of the pages (and passages thereof) are useful for assisting a human to fact-check that claim. There was also a second subtask, asking to determine the factuality of the claim} \cite{clef-checkthat-T2:2019}. The most effective approach for this task used textual entailment and external data~\cite{T2-Ghanem:2019}.

\subsection{\ct 2018}

\paragraph{Task~1$_{2018}$} \cite{clef-checkthat-T1:2018} was identical to Task~1$_{2019}$. The best approaches used \textit{pseudo-speeches} as a concatenation of all interventions by a debater~\cite{T1-Zuo:2018},
and represented the entries with embeddings, part-of-speech tags, and syntactic dependencies~\cite{T1-Hansen:2018}.

\paragraph{Task~2$_{2018}$.} \emph{Given a check-worthy claim in the form of a (transcribed) sentence, determine whether the claim is likely to be true, half-true, or false} \cite{clef-checkthat-T2:2018}. The best approach retrieved relevant information from the Web, and fed the claim with the most similar Web-retrieved text to a convolutional neural network~\cite{T1-Hansen:2018}.

\section{Description of the Tasks}
\label{sec:tasks}

\begin{figure}[t]
\centering
\includegraphics[width=\columnwidth]{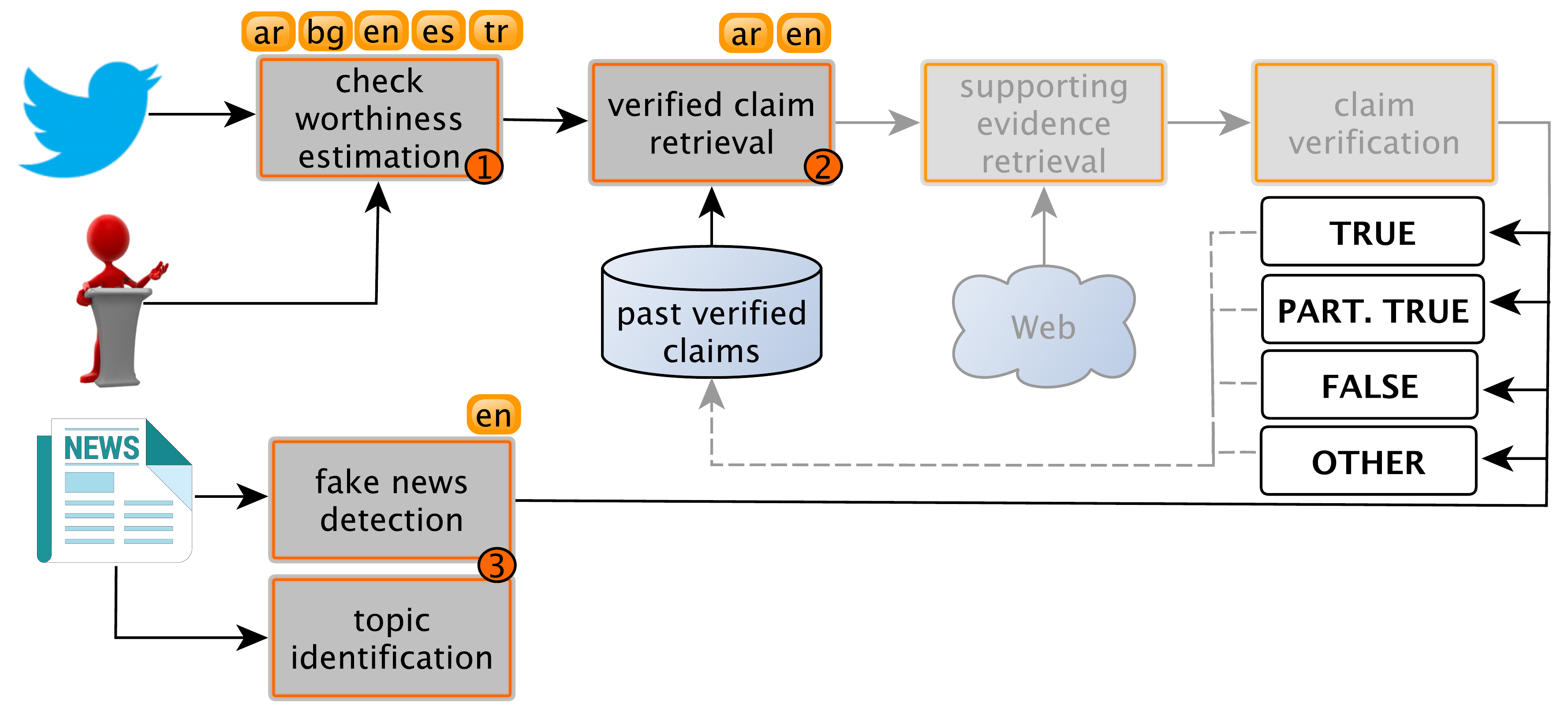}
\caption{The full verification pipeline. The 2021 lab covers three tasks from that pipeline: (\emph{i})~check-worthiness estimation, (\emph{ii})~verified claim retrieval, and (\emph{iii})~fake news detection. The gray tasks were addressed in previous editions of the lab~\cite{clef-checkthat-lncs:2020,clef-checkthat:2019}.
}
\label{fig:pipeline}
\end{figure}

The lab is organized around three tasks, each of which in turn has several subtasks. Figure~\ref{fig:pipeline} shows the full \ct verification pipeline, and the three tasks we target this year are highlighted.

\subsection{Task 1: Check-Worthiness Estimation} 
The aim of Task~1 is to determine whether a piece of text is worth fact-checking. In order to do that, we either resort to the judgments of professional fact-checkers or we ask human annotators to answer several auxiliary questions~\cite{alam2020call2arms,alam2020fighting}, such as ``does it contain a verifiable factual claim?'', ``is it harmful?'' and ``is it of general interest?'', before deciding on the final check-worthiness label.

\paragraph{\textbf{Subtask 1A: Check-worthiness of tweets.}} Given a tweet, produce a ranked list of tweets, ordered by their check-worthiness. This is a ranking task, focusing either on COVID-19 or politics. It was offered in Arabic, Bulgarian, English, Spanish, and Turkish. The participants were free to work on any language(s) of their choice, and they could also use multilingual approaches that make use of all datasets for training.

\paragraph{\textbf{Subtask 1B: Check-worthiness of debates or speeches.}} Given a political debate/speech, return a ranked list of its sentences, ordered by their check-worthiness. This is a ranking task, and it was offered in English.

\subsection{Task 2: Detecting Previously Fact-Checked Claims} 
Given a check-worthy claim in the form of a tweet, and a set of previously fact-checked claims, rank these previously fact-checked claims in order of their usefulness to fact-check that new claim.

\paragraph{\textbf{Subtask 2A: Detect previously fact-checked claims from tweets.}} Given a tweet, detect whether the claim it makes was previously fact-checked with respect to a collection of fact-checked claims. This is a ranking task, offered in Arabic and English, where the systems need to return a list of top-$n$ candidates.

\paragraph{\textbf{Subtask 2B: Detect previously fact-checked claims in political debates or speeches.}} Given a claim in a political debate or a speech, detect whether the claim has been previously fact-checked with respect to a collection of previously fact-checked claims. This is a ranking task, and it was offered in English.

\subsection{Task 3: Fake News Detection} 
Task~3 was offered for the first time, as a pilot task. In includes two subtasks.

\paragraph{\textbf{Subtask 3A: Multi-class fake news detection of news articles.}} Given the text of a news article, determine whether the claims made in the article are \emph{true}, \emph{partially true}, \emph{false}, or \emph{other}. This is a classification task, offered in English.

\paragraph{\textbf{Subtask 3B: Given the text of a news article, determine the topical domain of the article.}} This is a classification task to determine the topical domain of a news article~\cite{shahifakecovid}. It involves six categories (health, crime, climate, election, and education), and was offered in English.

\section{Datasets}
\label{sec:dataset}
 
Here, we briefly describe the datasets for each of the three tasks. For more details, refer to the task description paper for each individual task~\cite{clef-checkthat:2021:task2,clef-checkthat:2021:task1,clef-checkthat:2021:task3}.

\begin{table}[t]
\centering
\caption{\textbf{Task 1A (Check-worthiness in tweets):} Statistics about the CT--CWT--21
corpus for all five languages. The bottom part of the table shows the main topics.
}
\label{tab:datasets_task1}
\begin{tabular}{lcccccc}
\toprule
\bf Partition & \bf Arabic  & \bf Bulgarian & \bf English           & \bf Spanish   & \bf Turkish       & \bf Total \\ \midrule
Training    &   3,444   & 3,000     & \,\,\,\,822 & 2,495     & 1,899         & 11,660  \\
Development & \,\,\,\,661 & \,\,\,\,350 & \,\,\,\,140   & 1,247     &  \,\,\,\,388  & \,\,2,786\\
Testing & \,\,\,\,600 & \,\,\,\,357 & \,\,\,\,350   & 1,248     & 1,013         & \,\,3,568     \\
\hline
Total & 4,705 & 3,707 & 1,312 & 4,990 & 3,300 & 18,014\\
\\
\multicolumn{2}{l}{\bf Main topics}   \\  \midrule
COVID-19    & $\blacksquare$ & $\blacksquare$& $\blacksquare$ &               & $\blacksquare$ &\\
Politics    & $\blacksquare$ &               &               & $\blacksquare$ &   $\blacksquare$   & \\
\bottomrule
\end{tabular}
\end{table}

\subsection{Task 1: Check-Worthiness Estimation}

\paragraph{\textbf{Subtask 1A: Check-worthiness for tweets.}} 
We produced datasets in five languages with tweets covering COVID-19, politics, and other topics. We refer to these datasets as the CT--CWT--21 corpus, which stands for \ct check-worthiness for tweets 2021.
Table~\ref{tab:datasets_task1} shows statistics about the corpus. 

For \textbf{Arabic}, the training set is sampled from the corpus used in the 2020 edition of the \ct lab~\cite{clef-checkthat-ar:2020}; we only kept tweets with full agreement between the annotators. The tweets mainly cover politics and COVID-19. The newly collected testing set covers two political events: Gulf reconciliation and US Capitol riots.
They were labelled by two expert annotators, and the disagreements were resolved by discussion between the annotators.

For \textbf{Bulgarian}, we created a new dataset focusing on COVID-19. The tweets were annotated by three annotators, and disagreements were resolved by majority voting, and then by a consolidator.

For \textbf{English}, the dataset also focused on COVID-19. For training, we released the data used in the \ct lab of 2020~\cite{clef-checkthat-en:2020}.
For testing, we annotated new instances, where we had three annotators per example, and we resolved the disagreements by majority voting, and then by a consolidator.

For \textbf{Spanish}, we had a new dataset. The tweets were manually annotated by journalists from Newtral ---a Spanish fact-checking organization--- and came from the Twitter accounts of 300 Spanish politicians. 

For \textbf{Turkish}, the training set came from the TrClaim-19  dataset~\cite{kartal2020trclaim}, whereas the testing set was labelled for this task by three annotators. We applied majority voting for aggregation. The training set covers important events in Turkey in 2019 (e.g.,~the earthquake in Istanbul, and the military operation in Syria), whereas the test set focuses on COVID-19. 

The datasets for Arabic, Bulgarian, and English have annotations for some auxiliary questions. For example, annotators were asked question such as ``Is the claim of interest to the public?'' and ``Would the claim cause harm?''

\paragraph{\textbf{Subtask 1B: Check-worthiness for debates/speeches.}} 

For training, we collected 57 debates/speeches from 2012--2018, and we selected sentences from the transcript that were checked by human fact-checkers. After a political debate/speech, PolitiFact journalists publish an article fact-checking some of the claims made in it. We collected all such sentences and considered them check-worthy, and the rest non check-worthy. However, as PolitiFact journalists only fact-check a few claims made in the claims, there is an abundance of false negative examples in the dataset. To address this issue at test time, we manually looked over the debates from the test set and we attempted to check whether each sentence contains a verified claim using BM25 suggestions. Table~\ref{tab:datasets_task1b} shows some statistics about the data. Note the higher proportion of positive examples in the test set compared to the training and the development sets.

Further details about the CT--CWT--21 corpus for Task~1 can be found in~\cite{clef-checkthat:2021:task1}.

\begin{table}[tbh]
\centering
\caption{\textbf{Task 1B (Check-worthiness in Debates/Speeches):} Statistics about the CT--CWT--21 corpus for subtask 1B.}
\label{tab:datasets_task1b}
\begin{tabular}{@{}lrrr@{}}
\toprule
\multicolumn{1}{c}{\textbf{Dataset}} & \multicolumn{1}{c}{\textbf{\# of debates}} & \multicolumn{2}{c}{\textbf{\# of sentences}} \\
 & \multicolumn{1}{l}{} & \multicolumn{1}{l}{Check-worthy} & \multicolumn{1}{l}{Non-check-worthy} \\ \midrule
Training & 40 & 429 & 41,604 \\
Development & 9 & 69 & 3,517 \\
Test & 8 & 298 & 5,002 \\ \midrule
Total & 57 & 796 & 50,123 \\ \bottomrule
\end{tabular}
\end{table}

\subsection{Task 2: Detecting Previously Fact-Checked Claims}

\paragraph{\textbf{Subtask 2A: Detecting previously fact-checked claims from tweets.}}

For \textbf{English}, we have 1,401 annotated tweets, each matching a single claim in a set of 13,835 verified claims from Snopes.

For \textbf{Arabic}, we have 858 tweets, matching 1,039 verified claims (some tweets match more than one verified claim) in a collection of 30,329 previously fact-checked claims. The latter include 5,921 Arabic claims from AraFacts~\cite{ali2021arafacts} and 24,408 English claims from ClaimsKG~\cite{tchechmedjiev2019claimskg}, translated to Arabic using the Google translate API (\url{http://cloud.google.com/translate}).

\paragraph{\textbf{Subtask 2B: Detecting previously fact-checked claims in political debates/speeches.}} We have 669 claims from political debates \cite{shaar-etal-2020-known}, matched against 804 verified claims (some input claims match more than one verified claim) in a collection of 19,250 verified claims in PolitiFact.

Table~\ref{tab:datasets_task2} shows statistics about the CT--VCR--21 corpus for Task~2, including both subtasks and languages. CT--VCR--21 stands for \ct verified claim retrieval 2021. \emph{Input}--\emph{VerClaim} pairs represent input claims with their corresponding verified claims by a fact-checking source. The input for subtask 2A (2B) is a tweet (sentence from a political debate or a speech).
More details about the corpus construction can be found in~\cite{clef-checkthat:2021:task2}.

\begin{table}[t]
    \centering
    \caption{\textbf{Task 2:} Statistics about the CT--VCR--21 corpus,
    including the number of \emph{Input}--\emph{VerClaim}~pairs and the number of \emph{VerClaim}~claims to match the input claim against.}
    \label{tab:datasets_task2} 
    \begin{tabular}{l@{ }@{ }@{ }ccc}
        \toprule
                                  & \bf 2A--Arabic & \bf 2A--English & \bf 2B--English\\
        \midrule
        \bf Input claims          & \bf 858 & \bf 1,401 & \bf 669 \\
         \,\,\,\,\, Training     & 512 & \,\, 999 &  472 \\
         \,\,\,\,\, Development   & \,\, 85 & \,\, 200 &  119 \\
         \,\,\,\,\, Test          &  261 & \,\, 202 & \,\, 78 \\
        \midrule
        \bf Input--\emph{VerClaim}~pairs& \bf 1,039 & \bf 1,401 &\,\, \bf 804 \\
         \,\,\,\,\, Training    & \,\, 602 & \,\, 999 & \,\, 562 \\
         \,\,\,\,\, Development   & \,\, 102 & \,\, 200 & \,\, 139 \\
         \,\,\,\,\, Test          & \,\, 335 & \,\, 202 & \,\, 103 \\
        \midrule
        {\bf Verified claims} (to match against) & \bf 30,329 & \bf 13,835 & \bf 19,250\\
        \bottomrule
    \end{tabular}
\end{table}

\subsection{Task 3: Fake News Detection}

The process of corpus creation for Task~3 extends the AMUSED framework~\cite{shahi2020amused}. Starting with articles written by fact-checking organizations, we scraped the links to the original articles they verified, together with the factuality judgments. This process was done in two steps. First, in an automatic filtering step, all links with posts from social media channels or to multimedia documents were filtered out. In a second step, the remaining links were subjected to a manual checking process. During this step, we additionally made sure that the scraped link actually pointed to the checked document and that the document still existed (thus, eliminating error pages, articles with other content, etc.). After successful verification for each article, we scraped its title and full text.

\paragraph{\textbf{Subtask 3A: Multi-class fake news categorization of news articles.}} This subtask was offered in English only. We collected a total of 900 news articles for training and 354 news articles for testing from 11 fact-checking websites such as PolitiFact. The label for the original fact-checking site was given as a rating. However, due to the heterogeneous labeling schemes of different fact-checking organizations (e.g.,~\emph{false}: incorrect, inaccurate, misinformation; \emph{partially false}: mostly false, half false), we merged labels with shared meaning according to \cite{shahi2021exploratory}, resulting in the following four classes: \emph{false}, \emph{partially false}, \emph{true} and \emph{other}. We provided an ID, the title of the article, the text of the article, and our rating as data to the participants. No further metadata about the article was made available in the dataset. The ID is a unique identifier created for the dataset, the title is the title given in the target article, the text is the full-text content of the article, and our rating is the normalized rating provided in one of the above four label categories.

\paragraph{\textbf{Subtask 3B: Topical domain identification of news articles.}} 
This subtask is also offered in English only. We annotated a subset of the articles from subtask 3A with their topic: 318 articles for training, and 137 articles for testing in six different classes as shown in Table~\ref{d:3a} based on \cite{shahi_2021}. We refer to the corpus as CT-FAN-21, which stands for \ct 2021 Fake News. We provided the ID, the title, the text, and our rating as the metadata for the dataset. Here, ID is the unique ID, title is the title of the fake news article, the text is the full-text content of the article, and domain is the domain, expressed in terms of one of the above six categories.
  
The datasets for subtasks 3A and 3B are available in Zenodo \cite{gautam_kishore_shahi_2021_4714517}. We did not provide any other information (e.g.,~a link to the article, a publication date, eventual tags, authors, location of publication, etc.).
    
\begin{table}[t]
	\centering
	\caption{\textbf{Task 3:} Statistics about the number of documents and class distribution for the CT-FAN-21 corpus for fake news detection (left) and for topic identification (right).}
	\begin{tabular}{lcc}
		\toprule
		\bf Class    & \bf Training & \bf Test \\ \midrule
    	False & 465 & 111 \\ 
	    True & 142 & \,\,65  \\ 
		Partially false & 217 & 138 \\  
		Other & \,\,76 & \,\,40 \\  \midrule
		Total & 900 & 354 \\  \bottomrule
		\\\\
	\end{tabular}
	\hspace{15mm}
		\begin{tabular}{lcc}
		\toprule
		\textbf{Topic}   & \textbf{Training} & \textbf{Test} \\ \midrule
		Health & 127 & 54 \\ 
	    Climate & \,\,49 &  21   \\ 
		Economy & \,\,43 & 19 \\  
		Crime & \,\,39 & 17 \\ 
		Elections & \,\,32 & 14 \\  
		Education & \,\,28 & 12 \\  \midrule
		Total & 318 & 137 \\  \bottomrule
	\end{tabular}
	\label{d:3a}
\end{table}

\section{Evaluation}

For the ranking tasks, as in the two previous editions of the \ct lab, we used \emph{Mean Average Precision} (MAP) as the official evaluation measure. We further calculated and reported reciprocal rank, and $P@k$ for $k \in \{1, 3, 5, 10, 20, 30\}$, as unofficial measures. For the classification tasks, we used accuracy and macro-F$_1$ score. 

\section{Results for Task 1: Check-Worthiness Estimation}
\label{sec:task1}

Below, we report the evaluation results for task 1 and its two subtasks for all five languages.

\subsection{Task 1A. Check-Worthiness of Tweets}

Fifteen teams took part in this task, with English and Arabic being the most popular languages. Four out of the fifteen teams submitted runs for all five languages ---most of them having trained independent models for each language (yet, team UPV trained a single multilingual model).
For all five languages, we had a monolingual baseline based on $n$-gram representations.
Table~\ref{tab:results_task1a_all} shows the performance of the official submissions on the test set, in addition to the $n$-gram baseline. The official run was the last valid blind submission by each team. The table shows the runs ranked on the basis of the official MAP measure and includes all five languages.

\paragraph{\textbf{Arabic}}
Eight teams participated for Arabic, submitting a total of 17 runs (yet, recall that only the last submission counts). All participating teams fine-tuned existing pre-trained models, such as AraBERT, and multilingual BERT models. We can see that the top two systems additionally worked on improved training datasets. Team \textbf{Accenture} used a label augmentation approach to increase the number of positive examples, while team \textbf{bigIR} augmented the training set with the Turkish training set (which they automatically translated to Arabic).

\begin{table}
\centering
\caption{Task 1A: results for the official submissions in all five languages.}
\label{tab:results_task1a_all}
\begin{tabular}{@{}rlccccccccc@{}}
\toprule
& \bf Team & \bf MAP & \bf MRR & \bf RP & \bf P@1 & \bf P@3 & \bf P@5 & \bf P@10 & \bf P@20 & \bf P@30 \\ \midrule
\multicolumn{2}{l}{\bf Arabic}  \\
1 & Accenture~\cite{clef-checkthat:2021:task1:accenture} & 0.658 & 1.000 & 0.599 & 1.000 & 1.000 & 1.000 & 1.000 & 0.950 & 0.840 \\
2 & bigIR & 0.615 & 0.500 & 0.579 & 0.000 & 0.667 & 0.600 & 0.600 & 0.800 & 0.740 \\
3 & SCUoL \cite{clef-checkthat:2021:task1:althabiti2021}& 0.612 & 1.000 & 0.599 & 1.000 & 1.000 & 1.000 & 1.000 & 0.950 & 0.780 \\
4 & iCompass & 0.597 & 0.333 & 0.624 & 0.000 & 0.333 & 0.400 & 0.400 & 0.500 & 0.640 \\
4 & QMUL-SDS~\cite{clef-checkthat:2021:task1:S.Abumansour2021} & 0.597 & 0.500 & 0.603 & 0.000 & 0.667 & 0.600 & 0.700 & 0.650 & 0.720 \\
6 & TOBB ETU~\cite{clef-checkthat:2021:task1:tobbetu} & 0.575 & 0.333 & 0.574 & 0.000 & 0.333 & 0.400 & 0.400 & 0.500 & 0.680 \\
7 & DamascusTeam & 0.571 & 0.500 & 0.558 & 0.000 & 0.667 & 0.600 & 0.800 & 0.700 & 0.640 \\
8 & UPV~\cite{clef-checkthat:2021:task1:Schlicht2021} & 0.548 & 1.000 & 0.550 & 1.000 & 0.667 & 0.600 & 0.500 & 0.400 & 0.580 \\
9 & ngram-baseline & 0.428 & 0.500 & 0.409 & 0.000 & 0.667 & 0.600 & 0.500 & 0.450 & 0.440 \\ \midrule

\multicolumn{2}{l}{\bf Bulgarian}  \\
1 & bigIR & 0.737 & 1.000 & 0.632 & 1.000 & 1.000 & 1.000 & 1.000 & 1.000 & 0.800 \\
2 & UPV~\cite{clef-checkthat:2021:task1:Schlicht2021} & 0.673 & 1.000 & 0.605 & 1.000 & 1.000 & 1.000 & 1.000 & 0.800 & 0.700 \\
3 & ngram-baseline & 0.588 & 1.000 & 0.474 & 1.000 & 1.000 & 1.000 & 0.900 & 0.750 & 0.640 \\
4 & Accenture~\cite{clef-checkthat:2021:task1:accenture} & 0.497 & 1.000 & 0.474 & 1.000 & 1.000 & 0.800 & 0.700 & 0.600 & 0.440 \\
5 & TOBB ETU~\cite{clef-checkthat:2021:task1:tobbetu} & 0.149 & 0.143 & 0.039 & 0.000 & 0.000 & 0.000 & 0.200 & 0.100 & 0.060 \\ 
\midrule

\multicolumn{2}{l}{\bf English}  \\
\,\,1 & NLP\&IR@UNED~\cite{clef-checkthat:2021:task3:Martinez-Rico} & 0.224 & 1.000 & 0.211 & 1.000 & 0.667 & 0.400 & 0.300 & 0.200 & 0.160 \\
\,\,2 & Fight for 4230~\cite{clef-checkthat:2021:task1:Zhou2021} & 0.195 & 0.333 & 0.263 & 0.000 & 0.333 & 0.400 & 0.400 & 0.250 & 0.160 \\
\,\,3 & UPV~\cite{clef-checkthat:2021:task1:Schlicht2021} & 0.149 & 1.000 & 0.105 & 1.000 & 0.333 & 0.200 & 0.200 & 0.100 & 0.120 \\
\,\,4 & bigIR & 0.136 & 0.500 & 0.105 & 0.000 & 0.333 & 0.200 & 0.100 & 0.100 & 0.120 \\
\,\,5 & GPLSI~\cite{clef-checkthat:2021:task1:Sepulveda2021} & 0.132 & 0.167 & 0.158 & 0.000 & 0.000 & 0.000 & 0.200 & 0.150 & 0.140 \\
\,\,6 & csum112 & 0.126 & 0.250 & 0.158 & 0.000 & 0.000 & 0.200 & 0.200 & 0.150 & 0.160 \\
\,\,7 & abaruah & 0.121 & 0.200 & 0.158 & 0.000 & 0.000 & 0.200 & 0.200 & 0.200 & 0.140 \\
\,\,8 & NLytics~\cite{clef-checkthat:2021:task3:nlytics2021} & 0.111 & 0.071 & 0.053 & 0.000 & 0.000 & 0.000 & 0.000 & 0.050 & 0.120 \\
\,\,9 & Accenture~\cite{clef-checkthat:2021:task1:accenture} & 0.101 & 0.143 & 0.158 & 0.000 & 0.000 & 0.000 & 0.200 & 0.200 & 0.100 \\
10 & TOBB ETU~\cite{clef-checkthat:2021:task1:tobbetu} & 0.081 & 0.077 & 0.053 & 0.000 & 0.000 & 0.000 & 0.000 & 0.050 & 0.080 \\
11 & ngram-baseline & 0.052 & 0.020 & 0.000 & 0.000 & 0.000 & 0.000 & 0.000 & 0.000 & 0.020 \\
\midrule

\multicolumn{2}{l}{\bf Spanish}  \\
1 & TOBB ETU~\cite{clef-checkthat:2021:task1:tobbetu} & 0.537 & 1.000 & 0.525 & 1.000 & 1.000 & 0.800 & 0.900 & 0.700 & 0.680 \\
2 & GPLSI~\cite{clef-checkthat:2021:task1:Sepulveda2021} & 0.529 & 0.500 & 0.533 & 0.000 & 0.667 & 0.600 & 0.800 & 0.750 & 0.620 \\
3 & bigIR & 0.496 & 1.000 & 0.483 & 1.000 & 1.000 & 0.800 & 0.800 & 0.600 & 0.620 \\
4 & NLP\&IR@UNED~\cite{clef-checkthat:2021:task3:Martinez-Rico} & 0.492 & 1.000 & 0.475 & 1.000 & 1.000 & 1.000 & 0.800 & 0.800 & 0.620 \\
5 & Accenture~\cite{clef-checkthat:2021:task1:accenture} & 0.491 & 1.000 & 0.508 & 1.000 & 0.667 & 0.800 & 0.900 & 0.700 & 0.620 \\
6 & ngram-baseline & 0.450 & 1.000 & 0.450 & 1.000 & 0.667 & 0.800 & 0.700 & 0.700 & 0.660 \\
7 & UPV & 0.446 & 0.333 & 0.475 & 0.000 & 0.333 & 0.600 & 0.800 & 0.650 & 0.580\\
\midrule

\multicolumn{2}{l}{\bf Turkish}  \\
1 & TOBB ETU~\cite{clef-checkthat:2021:task1:tobbetu} & 0.581 & 1.000 & 0.585 & 1.000 & 1.000 & 0.800 & 0.700 & 0.750 & 0.660 \\
2 & SU-NLP~\cite{clef-checkthat:2021:task1:sunlp} & 0.574 & 1.000 & 0.585 & 1.000 & 1.000 & 1.000 & 0.800 & 0.650 & 0.680 \\
3 & bigIR & 0.525 & 1.000 & 0.503 & 1.000 & 1.000 & 1.000 & 0.800 & 0.700 & 0.720 \\
4 & UPV~\cite{clef-checkthat:2021:task1:Schlicht2021} & 0.517 & 1.000 & 0.508 & 1.000 & 1.000 & 1.000 & 1.000 & 0.850 & 0.700 \\
5 & Accenture~\cite{clef-checkthat:2021:task1:accenture} & 0.402 & 0.250 & 0.415 & 0.000 & 0.000 & 0.400 & 0.400 & 0.650 & 0.660 \\
6 & ngram-baseline & 0.354 & 1.000 & 0.311 & 1.000 & 0.667 & 0.600 & 0.700 & 0.600 & 0.460 \\

\bottomrule
\end{tabular}
\end{table}

\paragraph{\textbf{Bulgarian}}
Four teams took part for Bulgarian, submitting a total of 11 runs. The top-ranked team was \textbf{bigIR}. They did not submit a task description paper, and thus we cannot give much detail about their system. Team \textbf{UPV} is the second best system, and they used multilingual sentence transformer representation (SBERT) with knowledge distillation. They also introduced an auxiliary language identification task, aside from the downstream check-worthiness task.

\paragraph{\textbf{English}}
Ten teams took part in task 1A for English, with a total of 21 runs. The top-ranked team was \textbf{NLP\&IR@UNED}, and they fine-tuned several pre-trained transformers models. They reported BERTweet was best on the development set. The model was trained using RoBERTa on 850 million English tweets and 23 million COVID-19 related English tweets. The second best system (Team \textbf{Fight for 4230}) also used BERTweet with a dropout layer. It also included pre-processing and data augmentation.

\paragraph{\textbf{Spanish}}

Six teams took part for Spanish, with a total of 13 runs. 
The top team \textbf{TOBB ETU} explored different data augmentation strategies, including machine translation and weak supervision. However, they submitted a fine-tuned BETO model without any data augmentation. The first runner up \textbf{GPLSI} opted for using the BETO Spanish transformer together with a number of hand-crafted features, such as the presence of numbers or words in the LIWC lexicon.

\paragraph{\textbf{Turkish}}
Five teams participated for Turkish, submitting a total of 9 runs. All participants used BERT-based models. The top ranked team \textbf{TOBB ETU} fine-tuned BERTurk after removing user mentions and URLs. The runner up team \textbf{SU-NLP} applied a pre-processing step that includes removing hashtags, emojis, and replacing URLs and mentions with special tokens. Subsequently, they used an ensemble of BERTurk models fine-tuned with different seed values. The third-ranked team \textbf{bigIR} machine-translated the Turkish text to Arabic and then fine-tuned AraBERT on the translated text.

\paragraph{\textbf{All languages.}}
Table~\ref{tab:results_task1a_avg} summarizes the MAP performance of all the teams that submitted predictions for all languages in Task 1A. We can see that team \textbf{BigIR} performed best overall.

\begin{table}[h]
\centering
\caption{MAP performance for the official submissions to \textbf{Task 1A} in all five languages. $\mu$ shows a standard mean of the five MAP scores; $\mu_{w}$ shows a weighed mean, where each MAP is multiplied by the size of the testing set.}
\label{tab:results_task1a_avg}
\begin{tabular}{@{}rlccccccc@{}}
\toprule
 & \bf Team & \bf ar    & \bf bg    & \bf en & \bf es       & \bf tr    & \bf $\mu$ & $\mu_{w}$ \\ \midrule
1 & bigIR           & 0.615     & \bf 0.737 & 0.136     & 0.496     & 0.525     & \bf 0.502  & \bf 0.513  \\
2 & UPV~\cite{clef-checkthat:2021:task1:Schlicht2021}              & 0.548     & 0.673     & \bf 0.149 & 0.446     & 0.517    & 0.467      & 0.477 \\
3 & TOBB ETU~\cite{clef-checkthat:2021:task1:tobbetu}        & 0.575     & 0.149     & 0.081     & \bf 0.537 & \bf 0.581 & 0.385      & 0.472  \\
4 & Accenture~\cite{clef-checkthat:2021:task1:accenture}       & \bf 0.658 & 0.497     & 0.101     & 0.491     & 0.402     & 0.430      & 0.456  \\
5 & ngram-baseline & 0.428      & 0.588     & 0.052     & 0.450     & 0.354     & 0.374      & 0.394 \\ \bottomrule
\end{tabular}
\end{table}

\subsection{Task 1B. Check-Worthiness of Debates/Speeches}

Two teams took part in this subtask, submitting a total of 3 runs. Table~\ref{tab:results_task1b_english} shows the performance of the official submissions on the test set, in addition to the ngram baseline. Similarly to Task 1A, the official run was the last valid blind submission by each team. The table shows the runs ranked on the basis of the official MAP measure.

The top-ranked team, \textbf{Fight for 4230}, fine-tuned BERTweet after normalizing the claims, augmenting the data using WordNet-based substitutions and removal of punctuation. They were able to beat the ngram baseline by 18 MAP points absolute.

\begin{table}[t]
\centering
\caption{\textbf{Task 1B (English):} Official evaluation results, in terms of MAP, MRR, R-Precision, and Precision@$k$. The teams are ranked by the official evaluation measure: MAP.}
\label{tab:results_task1b_english}
\resizebox{\textwidth}{!}{
\begin{tabular}{@{}clccccccccc@{}}
\toprule
\bf Rank & \bf Team & \bf MAP & \bf MRR & \bf RP & \bf P@1 & \bf P@3 & \bf P@5 & \bf P@10 & \bf P@20 & \bf P@30 \\ \midrule
1 & Fight for 4230~\cite{clef-checkthat:2021:task1:Zhou2021} & 0.402 & 0.917 & 0.403 & 0.875 & 0.833 & 0.750 & 0.600 & 0.475 & 0.350 \\
2 & ngram-baseline & 0.235 & 0.792 & 0.263 & 0.625 & 0.583 & 0.500 & 0.400 & 0.331 & 0.217 \\
3 & NLytics~\cite{clef-checkthat:2021:task3:nlytics2021} & 0.135 & 0.345 & 0.130 & 0.250 & 0.125 & 0.100 & 0.137 & 0.156 & 0.135 \\ \bottomrule
\end{tabular}%
}
\end{table}

\FloatBarrier

\section{Results for Task 2: Verified Claim Retrieval}\label{sec:task2}

\subsection{Subtask 2A: Detecting Previously Fact-Checked Claims in Tweets}

Table~\ref{tab:results_task2a} shows the official results for Task 2A in both Arabic and English. A total of four teams participated in this task, and they all managed to improve over the Elastic Search (ES) baseline.

\paragraph{\textbf{Arabic}}
One team, bigIR, submitted a run for this subtask. They used AraBERT to rerank a list of candidates retrieved by a BM25 model. Their approach consists of three main steps. First, constructing a balanced training dataset, where the positive examples correspond to the query relevances (qrels) provided by the organizers, while the negative examples were selected from the top retrieved candidates by BM25 such that they were not already labeled as positive. Second, they fine-tuned AraBERT to predict the relevance score for a given tweet--VerClaim pair. They added two neural network layers on top of AraBERT to perform the classification task. Finally, at inference time, they first used BM25 to retrieve the top-20 candidate verified  claims. Then, they fed each tweet--VerClaim pair to the fine-tuned model to get a relevance score and to rerank the candidate claims accordingly. As Table~\ref{tab:results_task2a} shows, team \textbf{bigIR} outperformed the Elastic Search baseline by a good margin achieving a MAP@5 of 0.908 versus 0.794 for the baseline.

\paragraph{\textbf{English}} Three teams participated for English, submitting a total of ten runs. All of them managed to improve over the Elastic Search (ES) baseline by a large margin. Team \textbf{Aschern} had the top-ranked system, which used TF.IDF, fine-tuned pre-trained sentence-BERT, and the reranking LambdaMART model. The system is 13.4 (MAP@5) points absolute above the baseline. The second best system is the \textbf{NLytics}, which used RoBERTa to train their model and this system was 5 (MAP@5) point above the baseline.

\begin{table}[t]
\centering
\caption{\textbf{Task 2A:}
Official evaluation results, in terms of MRR, MAP@$k$, and Precision@$k$. The teams are ranked by the official evaluation measure: MAP@5. Here, \emph{ES-baseline} refers to the Elastic Search baseline.
\label{tab:results_task2a}}
\resizebox{\textwidth}{!}{%
\begin{tabular}{@{}clccccccccccc@{}}
\toprule
& \bf Team & \bf MRR & \multicolumn{5}{c}{\bf MAP} & \multicolumn{5}{c}{\bf Precision} \\ \midrule
 &  &  & \bf @1 & \bf @3 & \bf @5 & \bf @10 & \bf @20 & \bf @1 & \bf @3 & \bf @5 & \bf @10 & \bf @20 \\ \midrule

\multicolumn{2}{l}{\bf Arabic}  \\
1 & bigIR & 0.924 & 0.787 & 0.905 & \bf 0.908 & 0.910 & 0.912 & 0.908 & 0.391 & 0.237 & 0.120 & 0.061 \\
2 & ES-baseline & 0.835 & 0.682 & 0.782 & \bf 0.794 & 0.799 & 0.802 & 0.793 & 0.344 & 0.217 & 0.113 & 0.058 \\
 \midrule
 \multicolumn{2}{l}{\bf English}  \\
 1 & Aschern~\cite{clef-checkthat:2021:task3:Chernyavskiy2021} & 0.884 & 0.861 & 0.880 & \bf 0.883 & 0.884 & 0.884 & 0.861 & 0.300 & 0.182 & 0.092 & 0.046 \\
2 & NLytics~\cite{clef-checkthat:2021:task3:nlytics2021} & 0.807 & 0.738 & 0.792 & \bf 0.799 & 0.804 & 0.806 & 0.738 & 0.289 & 0.179 & 0.093 & 0.048 \\
3 & DIPS~\cite{clef-checkthat:2021:task2:DIPS} & 0.795 & 0.728 & 0.778 & \bf 0.787 & 0.791 & 0.794 & 0.728 & 0.282 & 0.177 & 0.092 & 0.048 \\
4 & ES-baseline & 0.761 & 0.703 & 0.741 & \bf 0.749 & 0.757 & 0.759 & 0.703 & 0.262 & 0.164 & 0.088 & 0.046 \\
 \bottomrule
\end{tabular}
}
\end{table}

\subsection{Subtask 2B: Detecting Previously Fact-Checked Claims in Political Debates and Speeches}

Table~\ref{tab:results_task2b} shows the official results for Task 2B, which was offered in English only. We can see that only three teams participated in this subtask, submitting a total of five runs, and no team managed to beat the Elastic Search (ES) baseline, which was based on BM25. 

Among the three participating teams, Team \textbf{DIPS} was the top-ranked one. They used sentence BERT (S-BERT) embeddings for all claims, and computed the cosine similarity for each pair of an input claim and a verified claim from the dataset of previously fact-checked claims. They made a prediction was made by passing a sorted list of cosine similarities to a neural network. Team \textbf{BeaSku} was the second-best team, which used a triplet loss training method to perform fine-tuning of the S-BERT model. Then, they used the scores predicted by the fine-tuned model along with BM25 scores as features to train a reranker  based on rankSVM. In addition, they discussed the impact of applying online mining of triplets. They also performed some experiments aiming at augmenting the training dataset with additional examples.

\begin{table}[tbh]
\centering
\caption{\textbf{Task 2B (English):}
Official evaluation results, in terms of MAP, MAP@$k$, and Precision@$k$. The teams are ranked by the official evaluation measure: MAP@5.
\label{tab:results_task2b}}

\resizebox{\textwidth}{!}{%
\begin{tabular}{clccccccccccccccc}
\toprule
 & \bf Team & \bf MRR & \multicolumn{5}{c}{\bf MAP} & \multicolumn{5}{c}{\bf Precision} \\ \midrule
 &  &  & \bf @1 & \bf @3 & \bf @5 & \bf @10 & \bf @20 & \bf @1 & \bf @3 & \bf @5 & \bf @10 & \bf @20 \\ \midrule
1 & ES-baseline & 0.350 & 0.304 & 0.339 & \bf 0.346 & 0.351 & 0.353 & 0.304 & 0.143 & 0.091 & 0.052 & 0.027 \\
2 & DIPS~\cite{clef-checkthat:2021:task2:DIPS} & 0.336 & 0.278 & 0.313 & \bf 0.328 & 0.338 & 0.342 & 0.266 & 0.143 & 0.099 & 0.059 & 0.032 \\
3 & Beasku~\cite{clef-checkthat:2021:task2:beasku2021} & 0.320 & 0.266 & 0.308 & \bf 0.327 & 0.332 & 0.332 & 0.253 & 0.139 & 0.101 & 0.056 & 0.028 \\
4 & NLytics~\cite{clef-checkthat:2021:task3:nlytics2021} & 0.216 & 0.171 & 0.210 & \bf 0.215 & 0.219 & 0.222 & 0.165 & 0.101 & 0.068 & 0.038 & 0.022 \\ \bottomrule
\end{tabular}%
}
\end{table}

\section{Overview of Task 3: Fake News Detection}
\label{sec:task3}

In this section, we present an overview of all task submissions for tasks 3A and 3B. Overall, there were 88 submissions by 27 teams for Task 3A and 49 submissions by 20 teams for task 3B. For task 3, unlike the other tasks, each participant could submit up to 5 runs. After evaluation, we found that two teams from task 3A and seven teams from task 3B submitted the wrong files, and thus we have not considered them for evaluation; we report the ranking for 25 teams for task 3A and 13 teams for task 3B. In Tables \ref{eval_results_task3A} and \ref{eval_results_task3B}, we report the best submission of each team for task 3A and 3B, respectively. In the following sections, we report the results for each of the subtasks.

\subsection{Task 3A. Multi-Class Fake News Detection of News Articles}
Most teams used deep learning models and in particular the transformer architecture for this pilot task. There have been no attempts to model knowledge with semantic technology, e.g.,~argument processing \cite{DumaniNS20}.

The best submission (team \textbf{NoFake}) was ahead of the rest by a rather large margin and achieved a macro-F1 score of 0.838. They applied BERT and made extensive use of external resources and in particular downloaded collections of misinformation datasets from fact-checking sites.
The second best submission (team \textbf{Saud}) achieved a macro-F1 score of 0.503 and used lexical features, traditional weighting methods as features, and standard machine learning algorithms. This shows, that traditional approaches can still outperform deep learning models for this task. Many teams used BERT and its newer variants. Such systems are ranked after the second position.
The most popular model was RoBERTa, which was used by seven teams. Team \textbf{MUCIC} used a majority voting ensemble with three BERT variants~\cite{clef-checkthat:2021:task3:MUCIC}.
The participating teams that used BERT had to find solutions for handling the length of the input: BERT and its variants have limitations on the length of their input, but the length of texts in the CT-FAN-21 dataset, which consists of newspaper articles, is much longer. In most cases, heuristics were used for the selection of part of the text. Overall, most submissions achieved a macro-F1 score below 0.5.

\begin{table}[t]
\caption{\textbf{Task 3A:} Performance of the best run per team based on F$_1$ score for individual classes, and accuracy and macro-F$_1$ for the overall measure. 
}
\begin{center}
\begin{tabular}{@{}cl|cccc|ccc@{}}
\toprule
& \textbf{Team} & 
\multicolumn{1}{p{.09\textwidth}}{\centering\textbf{True}} & 
\multicolumn{1}{p{.09\textwidth}}{\centering\textbf{False}} & 
\multicolumn{1}{p{.12\textwidth}}{\centering\textbf{Partially False}} & 
\multicolumn{1}{p{.09\textwidth}|}{\centering\textbf{Other}} & 
\textbf{Accuracy} & \textbf{Macro-F1} \\ \midrule
\,\,1 & NoFake*\cite{clef-checkthat:2021:task3:Kumari} & 0.824 & 0.862 & 0.879 & 0.785 & 0.853 & 0.838 \\  \,\,2 & Saud* & 0.321  & 0.615 & 0.502 & 0.618 & 0.537 & 0.514 \\ \,\,3 & DLRG* \cite{clef-checkthat:2021:task3:KannanDLRG} & 0.250 & 0.588 & 0.519 & 0.656 & 0.528 & 0.503 \\ \,\,4 & NLP\&IR@UNED \cite{clef-checkthat:2021:task3:Martinez-Rico}& 0.247 & 0.629 & 0.536 & 0.459 & 0.528 & 0.468 \\ 
\,\,5 & NITK\_NLP \cite{clef-checkthat:2021:task3:Hariharan}& 0.196 & 0.617 & 0.523 & 0.459 & 0.517 & 0.449 \\ 
\,\,6 & UAICS \cite{clef-checkthat:2021:task3:UAICS} & 0.442 & 0.470 & 0.482 & 0.391 & 0.458 & 0.446 \\ 
\,\,7 & CIVIC-UPM \cite{clef-checkthat:2021:task3:civicupm2021} & 0.268 & 0.577 & 0.472 & 0.340 & 0.463 & 0.414 \\ 
\,\,8 & Uni. Regensburg \cite{clef-checkthat:2021:task3:UnivRegensburg} & 0.231 & 0.489 & 0.497 & 0.400 & 0.438 & 0.404 \\ 
\,\,9 & Pathfinder* \cite{clef-checkthat:2021:task3:Tsoplefack} & 0.277 & 0.517 & 0.451 & 0.360 & 0.452 & 0.401 \\ 
10 & CIC* \cite{clef-checkthat:2021:task3:Ashraf} & 0.205 & 0.542 & 0.490 & 0.319 & 0.410 & 0.389 \\ 
11 & Black Ops
\cite{clef-checkthat:2021:task3:blackops2021} & 0.231 & 0.518 & 0.327 & 0.453 & 0.427 & 0.382 \\
12 & NLytics*  & 0.130 & 0.575 & 0.522 & 0.318 & 0.475 & 0.386 \\ 
13 & Nkovachevich \cite{clef-checkthat:2021:task3:Kovachevich}& 0.237 & 0.643 & 0.552 & 0.000 & 0.489 & 0.358 \\ 
14 & talhaanwar* & 0.283 & 0.407 & 0.435 & 0.301 & 0.367 & 0.357 \\ 
15 & abaruah & 0.165 & 0.531 & 0.552 & 0.125 & 0.455 & 0.343 \\ 
16 & Team GPLSI\cite{clef-checkthat:2021:task1:Sepulveda2021} & 0.293 & 0.602 & 0.226 & 0.092 & 0.356 & 0.303 \\ 
17 & Sigmoid \cite{clef-checkthat:2021:task3:sigmoid} & 0.222 & 0.345 & 0.323 & 0.154 & 0.291 & 0.261 \\ 
18 & architap & 0.154 & 0.291 & 0.394 & 0.187 & 0.294 & 0.257 \\ 
19 & MUCIC \cite{clef-checkthat:2021:task3:MUCIC} & 0.143 & 0.446 & 0.275 & 0.070 & 0.331 & 0.233 \\ 
20 & Probity & 0.163 & 0.401 & 0.335 & 0.033 & 0.302 & 0.233 \\ 
21 & M82B \cite{clef-checkthat:2021:task3:ashik} & 0.130 & 0.425 & 0.241 & 0.094 & 0.305 & 0.223 \\ 
22 & Spider & 0.046 & 0.482 & 0.145 & 0.069 & 0.316 & 0.186 \\ 
23 & Qword \cite{clef-checkthat:2021:task3:qword} & 0.108 & 0.458 & 0.000 & 0.033 & 0.277 & 0.150 \\ 
24 & ep* & 0.060 & 0.479 & 0.000 & 0.000 & 0.319 & 0.135 \\ 
25 & azaharudue* & 0.060 & 0.479 & 0.000 & 0.000 & 0.319 & 0.135 \\ 
\midrule
\multicolumn{2}{l|}{~~Majority class baseline} & 0.000 & 0.477 & 0.000 & 0.000 & 0.314 & 0.119\\

\bottomrule
\multicolumn{8}{l}{\footnotesize{* Runs submitted after the deadline, but before the release of the results.}}
\end{tabular}
\end{center}
\label{eval_results_task3A}
\end{table}

The second most popular neural network model was the recurrent neural network, which was used by six teams. Many participants experimented also with traditional text processing methods as they were commonly used for knowledge representation in information retrieval. For example, team \textbf{Kovachevich} used a Na\"{i}ve Bayes classifier with TF.IDF features for the 500 most frequent stems in the dataset~\cite{clef-checkthat:2021:task3:Kovachevich}.
Some lower-ranked teams used additional techniques and resources. 
These include LIWC~\cite{clef-checkthat:2021:task3:Martinez-Rico}, data augmentation by inserting artificially created similar documents~\cite{clef-checkthat:2021:task3:Ashraf}, semantic analysis with the Stanford Empath Tool~\cite{clef-checkthat:2021:task3:UAICS}, and the reputation of the sites of a search engine result after searching with the title of the article~\cite{clef-checkthat:2021:task3:Martinez-Rico}.

\subsection{Task 3B. Topical Domain Identification of News Articles}
The performance of the systems for task 3B was overall higher than for task 3A. The first three submissions were close together and all used transformer-based architectures. The best submissionm, by team \textbf{NITK\_NLP}, used an ensemble of three transformers~\cite{clef-checkthat:2021:task3:Hariharan}. The second best submission (by team \textbf{NoFake}) and the third best submission (by team \textbf{Nkovachevich}) used BERT.

\begin{table}[t]
\centering
\caption{\textbf{Task 3B:} Performance of the best run per team based on F1-measure for individual classes, and accuracy and macro-F$_1$ for overall measure.}
\resizebox{\textwidth}{!}{%
\begin{tabular}{cl|cccccc|cc}
\toprule
& \textbf{Team} & \textbf{Climate} & \textbf{Crime} & \textbf{Economy} & \textbf{Education} & \textbf{Elections} & \textbf{Health} & \textbf{Acc} & \textbf{Macro F1} \\ \midrule
\,\,1 & NITK\_NLP \cite{clef-checkthat:2021:task3:Hariharan} & 0.950 & 0.872 & 0.824 & 0.800 & 0.897 & 0.946 & 0.905 & 0.881 \\ 
\,\,2 & NoFake*  & 0.800 & 0.875 & 0.900 & 0.957 & 0.692 & 0.907 & 0.869 & 0.855 \\ 
\,\,3 & Nkovachevich \cite{clef-checkthat:2021:task3:Kovachevich} & 0.927 & 0.872 & 0.743 & 0.737 & 0.857 & 0.911 & 0.869 & 0.841 \\ 
\,\,4 & DLRG & 0.952 & 0.743 & 0.688 & 0.800 & 0.828 & 0.897 & 0.847 & 0.818 \\
\,\,5 & CIC* \cite{clef-checkthat:2021:task3:Ashraf} & 0.952 & 0.750 & 0.688 & 0.588 & 0.889 & 0.871 & 0.832 & 0.790 \\ 
\,\,6 & architap & 0.900 & 0.711 & 0.774 & 0.609 & 0.815 & 0.907 & 0.825 & 0.786 \\ 
\,\,7 & NLytics & 0.826 & 0.714 & 0.710 & 0.500 & 0.769 & 0.867 & 0.788 & 0.731 \\
\,\,8 & CIVIC-UPM* \cite{clef-checkthat:2021:task3:civicupm2021} & 0.864 & 0.700 & 0.645 & 0.421 & 0.609 & 0.821 & 0.745 & 0.677 \\ 
\,\,9 & ep* & 0.727 & 0.476 & 0.222 & 0.343 & 0.545 & 0.561 & 0.511 & 0.479 \\ 
10 & Pathfinder* \cite{clef-checkthat:2021:task3:Tsoplefack}  & 0.900 & 0.348 & 0.250 & 0.000 & 0.526 & 0.667 & 0.599 & 0.448 \\ 
11 & M82B \cite{clef-checkthat:2021:task3:ashik} & 0.294 & 0.000 & 0.000 & 0.000 & 0.000 & 0.576 & 0.409 & 0.145 \\ 
12 & MUCIC \cite{clef-checkthat:2021:task3:MUCIC} & 0.294 & 0.000 & 0.000 & 0.000 & 0.000 & 0.576 & 0.409 & 0.145 \\ 
13 & azaharudue* & 0.129 & 0.000 & 0.000 & 0.125 & 0.000 & 0.516 & 0.321 & 0.128 \\ 
\midrule
\multicolumn{2}{c|}{\textit{Majority class baseline}} & 0.000 & 0.000 & 0.000 & 0.000 & 0.000 & 0.565 & 0.394 & 0.094 \\

\bottomrule
\multicolumn{10}{l}{\footnotesize{* Runs submitted after the deadline, but before the release of the results.}}
\end{tabular}
}
\label{eval_results_task3B}
\end{table}

\section{Related Work}
\label{sec:related}

There has been work on checking the factuality/credibility of a claim, of a news article, or of an information source~\cite{ba2016vera,baly-etal-2020-written,R17-1046,ma2016detecting,mukherjee2015leveraging,FANG,popat2016credibility,zubiaga2016analysing}. Claims can come from different sources, but special attention has been paid to those from social media~\cite{gupta2014tweetcred,mitra2015credbank,nakov2021automated,shaar2021role,shaar-etal-2020-known,shu2017fake,zhao2015enquiring}. 
Check-worthiness estimation is still a fairly-new problem especially in the context of social media~\cite{RANLP2017:debates,Hassan:15,Hassan2016ComparingAF,hassan2017claimbuster}. A lot of research was performed on fake news detection for news articles, which is mostly approached as a binary classification problem~\cite{oshikawa-etal-2020-survey}. 

\ct is related to several other initiatives at SemEval on determining rumour veracity and support for rumours \cite{derczynski2017semeval,gorrell2019semeval},
on stance detection \cite{mohammad2016semeval},
on fact-checking in community question answering forums \cite{mihaylova2019semeval}, 
on propaganda detection \cite{da2020semeval,dimitrov2021semeval2021},
and on semantic textual similarity \cite{agirre-etal-2016-semeval,nakov-EtAl:2016:SemEval}.
It is also related to the FEVER task~\cite{thorne-etal-2018-fever} on fact extraction and verification, as well as to the Fake News Challenge~\cite{hanselowski-etal-2018-retrospective}, and the FakeNews task at MediaEval~\cite{pogorelov2020fakenews}.

\newpage
\section{Conclusion and Future Work}
\label{sec:final}

We have presented the 2021 edition of the \ct Lab, which was the most popular CLEF-2021 lab in terms of team registrations (132 teams registered), and about one-third of them actually participated: 15, 5, and 25 teams submitted official runs for tasks 1, 2, and 3, respectively. The lab featured tasks that span important steps of the verification pipeline: from spotting check-worthy claims to checking whether they have been fact-checked elsewhere before. We further featured a fake news detection task, and we also checked the class and the topical domain of news articles. Together, these tasks support the technology pipeline to assist human fact-checkers. Moreover, in-line with the general mission of CLEF, we promoted multi-linguality by offering our tasks in five different languages.

In future work, we plan to extend the datasets with more examples, more information sources, and also to cover more languages.

\section*{Acknowledgments}
The work of Tamer Elsayed and Maram Hasanain is made possible by NPRP grant \#NPRP-11S-1204-170060 from the Qatar National Research Fund (a member of Qatar Foundation). The work of Fatima Haouari is supported by GSRA grant \#GSRA6-1-0611-19074 from the Qatar National Research Fund. The statements made herein are solely the responsibility of the authors. 

This research is also part of the Tanbih mega-project, developed at the Qatar Computing Research Institute, HBKU, which aims to limit the impact of ``fake news'', propaganda, and media bias, thus promoting digital literacy and critical thinking.

\bibliography{bib/sigproc,bib/clef20_checkthat,bib/clef19_checkthat,bib/clef18_checkthat,bib/clef21_checkthat} 
\bibliographystyle{splncs04}

\newpage
\clearpage
\section*{Appendix}
\label{sec:appendix}
\appendix
\section{Systems for Task 1}
\label{app:task1}
The positions in the task ranking appear after each team name. See \mbox{Tables~\ref{tab:results_task1a_all}--\ref{tab:results_task1b_english}} for further details.

\noindent
\textbf{Team Accenture~\cite{clef-checkthat:2021:task1:accenture} (\rank{1A:ar}{1} \rank{1A:bg}{4} \rank{1A:en}{9} \rank{1A:es}{5} \rank{1A:tr}{5})} used BERT and RoBERTa with data augmentation. They further generated additional synthetic training data using lexical substitution. To find the most probable substitutions, they used BERT-based contextual embedding to create synthetic examples for the positive class. They further added a mean-pooling layer and a dropout layer on top of the model before the final classification layer. 

\noindent
\textbf{Team Fight for 4230~\cite{clef-checkthat:2021:task1:Zhou2021} (\rank{1A:en}{2} \rank{1B:en}{1})} focused its efforts mostly on two fronts: the creation of a pre-processing module able to properly normalize the tweets and the augmentation of the data by means of machine translation and WordNet-based substitutions. The pre-processing included link removal and punctuation cleaning, as well as quantities and contractions expansion. All hashtags related to COVID-19 were normalized into one and the hashtags were expanded. Their best approach was based on BERTweet with a dropout layer and the above-mentioned pre-processing.

\noindent
\textbf{Team GPLSI~\cite{clef-checkthat:2021:task1:Sepulveda2021} (\rank{1A:en}{5} \rank{1A:es}{2}) } applied the RoBERTa and the BETO transformers together with different manually engineered features, such as the occurrence of dates and numbers or words from LIWC. A thorough exploration of parameters was made using weighting and bias techniques. They also tried to split the four-way classification into two binary classifications and one three-way classification. They further tried oversampling and undersampling.

\noindent\textbf{Team iCompass (\rank{ar}{4}) } used several prepossessing steps, including (\emph{i})~English word removal, (\emph{ii})~removing URLs and mentions, and (\emph{iii})~data normalization, removing tashkeel and the letter madda from texts, as well as duplicates, and replacing some characters to prevent mixing. They proposed a simple ensemble of two BERT-based models, which include AraBERT and Arabic-ALBERT.

\noindent\textbf{Team NLP\&IR@UNED~\cite{clef-checkthat:2021:task3:Martinez-Rico} (\rank{1A:en}{1} \rank{1A:es}{4}}) used several transformer models, such as BERT, ALBERT, RoBERTa, DistilBERT, and Funnel-Transformer, for the experiments to compare the performance. For English, they obtained better results using BERT trained with tweets. For Spanish, they used Electra. 

\noindent\textbf{Team NLytics~\cite{clef-checkthat:2021:task1:nlytics2021} (\rank{1A:en}{8} \rank{1B:en}{3})} used RoBERTa with a regression function in the final layer, approaching the problem as a ranking task. 

\noindent
\textbf{Team QMUL-SDS~\cite{clef-checkthat:2021:task1:S.Abumansour2021} (\rank{1A:ar}{4}) }
used the AraBERT preprocessing function to (\emph{i})~replace URLs, email addressees, and user mentions with standard words, (\emph{ii})~removed line breaks, HTML markup, repeated characters, and unwanted characters, such as emotion icons, and (\emph{iii})~handled white spaces between words and digits (non-Arabic, or English), and/or a combination of both, and before and after two brackets, and also (\emph{iv})~removed unnecessary punctuation. They addressed the task as a ranking problem, and fine-tuned an Arabic transformer (AraBERTv0.2-base) on a combination of the data from this year and the data from the CheckThat! lab 2020 (the CT20-AR dataset).

\noindent\textbf{Team SCUoL~\cite{clef-checkthat:2021:task1:althabiti2021} (\rank{1A:ar}{3}}) used typical pre-processing steps, including cleaning the text, segmentation, and tokenization. Their experiments consists of fine-tuning different AraBERT models, and their final results were obtained using AraBERTv2-base.

\noindent
\textbf{Team SU-NLP~\cite{clef-checkthat:2021:task1:sunlp} (\rank{1A:tr}{2}) } also used several pre-possessing steps, including (\emph{i})~removing emojis, hashtags, and (\emph{ii})~replacing all mentions with a special token (@USER), and all URLs with the respective website's domain. If the URL is for a tweet, they replaced the URL with TWITTER and the respective user account name. They reported that this URL expansion method improved the performance. Subsequently, they used an ensemble of BERTurk models fine-tuned using different seed values.

\noindent
\textbf{Team TOBB ETU~\cite{clef-checkthat:2021:task1:tobbetu} (\rank{1A:ar}{6} \rank{1A:bg}{5} \rank{1A:en}{10} \rank{1A:es}{1} \rank{1A:tr}{1}) } investigated different approaches to fine-tune transformer models including data augmentation using machine translation, weak supervision, and cross-lingual training. For their submission, they removed URLs and user mentions from the tweets, and fine-tuned a separate BERT-based models for each language. In particular, they fine-tuned BERTurk\footnote{http://huggingface.co/dbmdz/bert-base-turkish-cased}, AraBERT, BETO\footnote{http://huggingface.co/dccuchile/bert-base-spanish-wwm-cased}, and the BERT-base model for Turkish, Arabic, Spanish, and English, respectively. For Bulgarian, they fine-tune a RoBERTa model pre-trained with Bulgarian documents.\footnote{http://huggingface.co/iarfmoose/roberta-base-bulgarian}

\noindent 
\textbf{Team UPV~\cite{clef-checkthat:2021:task1:Schlicht2021} (\rank{1A:ar}{8} \rank{1A:bg}{2} \rank{1A:en}{3} \rank{1A:es}{6} \rank{1A:tr}{4}) } used a multilingual sentence transformer representation (S-BERT) with knowledge distillation, originally intended for question answering. They further introduced an auxiliary language identification task, aside the downstream check-worthiness task. 

\section{Systems for Task 2}
\label{app:task2}

\noindent\textbf{Team Aschern~\cite{clef-checkthat:2021:task3:Chernyavskiy2021} (\rank{2A:en}{1})} used TF.IDF, fine-tuned pre-trained S-BERT, and the reranking LambdaMART model.

\noindent
\textbf{Team BeaSku~\cite{clef-checkthat:2021:task2:beasku2021} (\rank{2B:en}{3})} used triplet loss training to fine-tune S-BERT. Then, they used the scores predicted by the fine-tuned model along with BM25 scores as features to train a rankSVM re-ranker. They further discussed the impact of applying online mining of triplets. They also experimented with data augmentation.

\noindent\textbf{Team DIPS~\cite{clef-checkthat:2021:task2:DIPS} (\rank{2A:en}{3} \rank{2B:en}{2}) } calculated S-BERT embeddings for all claims, then computed a cosine similarity for each pair of an input claim and a verified claim. The prediction is made by passing a sorted list of cosine similarities to a neural network.

\noindent\textbf{Team NLytics (\rank{2A:en}{2} \rank{2B:en}{4}) } approached the problem as a regression task, and used RoBERTa with a regression function in the final layer.

\section{Systems for Task 3}
\label{app:task3}

\noindent
\textbf{Team Black Ops~\cite{clef-checkthat:2021:task3:blackops2021} (\rank{3A}{11})} performed data pre-processing by removing stop-words and punctuation marks. Then, they experimented with decision trees, random forest, and gradient boosting classifiers for Task 3A, and found the latter to perform best. 

\noindent\textbf{Team CIC \cite{clef-checkthat:2021:task3:Ashraf} (\rank{3A}{10} \rank{3B}{5})} experimented with logistic regression, multi-layer perceptron, support vector machines, and random forest. Their experiments consisted of using stratified 5-fold cross-validation on the training data. Their best results were obtained using logistic regression for task 3A, and a multi-layer perceptron for task 3B.

\noindent\textbf{Team CIC \rank{3A}{11} } experimented with a decision tree, a random forest, and a gradient boosting algorithms. They found the latter to perform best.

\noindent\textbf{Team CIVIC-UPM~\cite{clef-checkthat:2021:task3:civicupm2021} (\rank{3A}{7} \rank{3B}{8})} participated in the two subtasks of task 3. They performed pre-processing, using a number of tools: (\emph{i})~\texttt{ftfy} to repair Unicode and emoji errors, (\emph{ii})~\texttt{ekphrasis} to perform lower-casing, normalizing percentages, time, dates, emails, phones, and numbers, (\emph{iii})~\texttt{contractions} for abbreviation expansion, and (\emph{iv})~\texttt{NLTK} for word tokenization, stop-words removal, punctuation removal and word lemmatization. Then, they combined \texttt{doc2vec} with transformer representations (Electra base, T5 small and T5 base, Longformer base, RoBERTa base and DistilRoBERTa base). 
They further used additional data from Kaggle’s Ag News task, Kaggle's KDD2020, and Clickbait news detection competitions. Finally, they experimented with a number of classifiers such as Na\"{i}ve Bayes, Random Forest, Logistic Regression with L1 and L2 regularization, Elastic Net, and SVMs. The best system for subtask 3A used DistilRoBERTa-base on the text body with oversampling and a sliding window for dealing with long texts.
Their best system for task 3B used RoBERTa-base on the title+body text with oversampling but no sliding window.

\noindent \textbf{Team DLRG (\rank{3A}{3} \rank{3B}{4})} experimented with a number of traditional approaches like Random Forest, Na\"{i}ve Bayes and Logistic Regression as well as an online passive-aggressive classifier and different ensembles thereof. The best result was achieved by an ensemble of Na\"{i}ve Bayes, Logistic Regression, and the Passive Aggressive classifier for task 3A. For task 3B, the Online Passive-Aggressive classifier outperformed all other approaches, including the considered ensembles.

\noindent
\textbf{Team GPLSI~\cite{clef-checkthat:2021:task1:Sepulveda2021} (\rank{3A}{16})} applied the RoBERTa transformer together with different manually-engineered features, such as the occurrence of dates and numbers or words from LIWC. Both the title and the body were concatenated as a single sequence of words. Rather than going for a single multi-class setting, they used two binary models considering the most frequent classes: false vs. other, and true vs. other, followed by one three-class model.

\noindent\textbf{Team MUCIC \cite{clef-checkthat:2021:task3:MUCIC} (\rank{3A}{19} \rank{3B}{12})} used a majority voting ensemble with three BERT variants. They applied BERT, Distilbert, and RoBERTa, and fine-tuned the pre-trained models.

\noindent\textbf{Team NITK\_NLP\cite{clef-checkthat:2021:task3:Hariharan} (\rank{3A}{5} \rank{3B}{1})} proposed an approach, that included pre-processing and tokenization of the news article, and then experimented with multiple transformer models. The final prediction was made by an ensemble.

\noindent\textbf{Team NKovachevich \cite{clef-checkthat:2021:task3:Kovachevich} (\rank{3A}{13} \rank{3B}{3})} created lexical features. They extracted the 500 most frequent word stems in the dataset, and calculated the TF.IDF values, which they used in a multinomial Naïve Bayes classifier. A much better performance was achieved with an LSTM model that used GloVe embeddings. A little lower F1 value was achieved using BERT. 
They further found RoBERTa to perform worse than BERT.

\noindent\textbf{Team NLP\&IR@UNED \cite{clef-checkthat:2021:task3:Martinez-Rico} (\rank{3A}{4})} experimented with four transformer architectures and input sizes of 150 and 200 words. In the preliminary tests, the best performance was achieved by ALBERT with 200 words. They also experimented with combining TF.IDF values from the text, all the features provided by the LIWC tool, and the TF.IDF values from the first 20 domain names returned by a query to a search engine. Unlike what was obtained in the dev dataset, in the official competition, the best results were obtained with the approach based on TF.IDF, LIWC, and domain names.

\noindent\textbf{Team NLytics (\rank{3A}{12} \rank{3B}{7})} fined-tuned RoBERTa on the dataset for each of the sub-tasks. Since the data is unbalanced, they used under-sampling. They also truncated the documents to 512 words to fit into the RoBERTa input size.

\noindent\textbf{Team NoFake \cite{clef-checkthat:2021:task3:Kumari} (\rank{3A}{1} \rank{3B}{2})}  applied BERT without fine-tuning, but used an extensive amount of additional data for training, downloaded from various fact-checking websites.

\noindent\textbf{Team Pathfinder \cite{clef-checkthat:2021:task3:Tsoplefack} (\rank{3A}{9} \rank{3A}{10}) } participated in both tasks and used multinomial Na\"{i}ve Bayes and random forest. The former performed better for both tasks. For task 3A, the they merged the classed \emph{false} and \emph{partially false} into one class, which boosted the model performance by 41\% (a non-official score mentioned in the paper).

\noindent
\textbf{Team Probity (\rank{3A}{20})} addressed the multiclass fake news detection subtask, they used a simple LSTM architecture where they adopted word2vec embeddings to represent the news articles.

\noindent\textbf{Team Qword \cite{clef-checkthat:2021:task3:qword} (\rank{3A}{23}) } applied pre-processing techniques, which included stop-word removal, punctuation removal and lemmatization using a Porter stemmer. The TF.IDF values were calculated for the words. For these features, four classification algorithms were applied. The best result was given by Extreme Gradient Boosting. 

\noindent
\textbf{Team SAUD (\rank{3A}{2})} 
used an SVM with TF.IDF. They tried Logistic Regression, Multinomial Na\"{i}ve Bayes, and Random Forest, and found SVM to work best.  

\noindent\textbf{Team Sigmoid \cite{clef-checkthat:2021:task3:sigmoid} (\rank{3A}{17})} experimented with different traditional machine learning approaches, with multinomial Na\"{i}ve Bayes performing best, and one deep learning approach, namely an LSTM with the Adam optimizer. The latter outperformed the more traditional approaches.

\noindent\textbf{Team Spider (\rank{3A}{22}) } applies an LSTM, after a pre-processing consisting of stop-word removal and stemming. 

\noindent
\textbf{Team UAICS~\cite{clef-checkthat:2021:task3:UAICS} (\rank{3A}{6})} experimented with various models including BERT, LSTM, Bi-LSTM, and feature-based models. Their submitted model is a Gradient Boosting with a weighted combination of three feature groups: bi-grams, POS tags, and lexical categories of words.

\noindent\textbf{Team University of Regensburg \cite{clef-checkthat:2021:task3:UnivRegensburg} (\rank{3A}{8})} used different fine-tuned variants of BERT with a linear layer on top and applied different approaches to address the maximum sequence length of BERT. Besides hierarchical transformer representations, they also experimented with different summarization techniques like extractive and abstractive summarization. They performed oversampling to address the class imbalance, as well as extractive (using DistilBERT) and abstractive summarization (using distil-BART-CNN-12-6), before performing classification using fine-tuned BERT with a hierarchical transformer representation.

\end{document}